\documentclass[fleqn,10pt]{wlscirep}
\usepackage[utf8]{inputenc}
\usepackage[T1]{fontenc}
\usepackage{caption}
\usepackage{subcaption}
\usepackage{xspace}
\usepackage{multirow}

\newcommand{\ie}{i.e.\@\xspace} 

\newcommand{\titleDataset}{MedFMC\@\xspace}
\title{MedFMC: A Real-world Dataset and Benchmark For Foundation Model Adaptation in Medical Image Classification}

\author[1,2,*]{Dequan Wang}
\author[1,*]{Xiaosong Wang}
\author[1]{Lilong Wang}
\author[1]{Mengzhang Li}
\author[3]{Qian Da}  
\author[4]{Xiaoqiang Liu} 
\author[5]{Xiangyu Gao} 
\author[6]{Jun Shen} 
\author[1]{Junjun He}
\author[7]{Tian Shen}
\author[7]{Qi Duan}
\author[8]{Jie Zhao}
\author[1,9]{Kang Li} 
\author[1,$\dag$]{Yu Qiao}
\author[1,$\dag$]{Shaoting Zhang}
\affil[1]{Shanghai AI Laboratory, China}
\affil[2]{Shanghai Jiaotong University, China}
\affil[3]{Shanghai Ruijing Hospital, School of Medicine, Shanghai Jiao Tong University, China}
\affil[4]{Shanghai Tenth People's Hospital of Tongji University, China}
\affil[5]{Xuzhou Central Hospital, China}
\affil[6]{Renji Hospital, School of Medicine, Shanghai Jiao Tong University, China}
\affil[7]{Sensetime Research, China}
\affil[8]{The First Affiliated Hospital of Zhengzhou University, China}
\affil[9]{West China Hospital, Sichuan University, China}

\affil[*]{These authors contributed equally to this work}
\affil[$\dag$]{Corresponding author(s): qiaoyu@pjlab.org.cn, zhangshaoting@pjlab.org.cn}

\begin{abstract}

Foundation models, often pre-trained with large-scale data, have achieved paramount success in jump-starting various vision and language applications. 
Recent advances further enable adapting foundation models in downstream tasks efficiently using only a few training samples, e.g., in-context learning.
Yet, the application of such learning paradigms in medical image analysis remains scarce due to the shortage of publicly accessible data and benchmarks. 
In this paper, we aim at approaches adapting the foundation models for medical image classification and present a novel dataset and benchmark for the evaluation, \ie, examining the overall performance of accommodating the large-scale foundation models downstream on a set of diverse real-world clinical tasks.
We collect five sets of medical imaging data from multiple institutes targeting a variety of real-world clinical tasks (22,349 images in total), \ie, thoracic diseases screening in X-rays, pathological lesion tissue screening, lesion detection in endoscopy images, neonatal jaundice evaluation, and diabetic retinopathy grading. 
Results of multiple baseline methods are demonstrated using the proposed dataset from both accuracy and cost-effective perspectives.

\end{abstract}
\begin{document}

\flushbottom
\maketitle
\thispagestyle{empty}


\section*{Background \& Summary}
In the new trend of training even larger and universal foundation models (e.g., Vision Transformers~\cite{dosovitskiyimage}, GPTs~\cite{radford2018improving}, PubmedBERT~\cite{pubmedbert}, and CLIP~\cite{radford2021learning}) using thousands of millions of data samples (sometimes in multiple modalities), developing cost-effective model adaptation methods for detailed applications become the new gold, especially when it only demands very few data samples. On the other side, the shortage of publicly accessible datasets in medical imaging has largely blocked the development and application of large-scale deep learning models (training from scratch) in many clinical downstream tasks. It is because obtaining quality annotations remains a tedious task for medical professionals, e.g., hand-label volumetric data repeatedly. Providing a few textbook sample cases is more logically feasible and complies with the training process of medical residents.  In the domain of medical image analysis, it is even more valuable to promote such learning paradigms when diseased cases are often rare in comparison to the numerous amount of normal population. 

The common fine-tuning scheme~\cite{shin2016deep} with ImageNet~\cite{deng2009imagenet} pre-trained models can diminish the need of large-scale data for the train-from-scratch scheme. However, it still requires a fair amount of data for faster fine-tuning while avoiding overfitting. Alternatively, few-shot methods could leverage more on the distinctive representation produced by the foundation models, which has succeeded in considerable language modeling~\cite{brown2020language} and vision~\cite{dhillon2019baseline,tian2020rethinking} tasks. The existing techniques of adapting foundation models in medical image analysis~\cite{ouyang2020self,singh2021metamed} demand the employment of dedicated medical pre-trained models that is hard to produce even if self-supervised learning is utilized. Recently, cutting-edge techniques, e.g., prompt-based learning~\cite{zhou2022learning,zhou2022conditional}, can leverage the foundation models pre-trained (via self-supervised learning, e.g., DINO~\cite{caron2021emerging} and MAE~\cite{he2022masked}) using vast amounts of data from multiple modalities and domains and transfer these universal representations to tasks with very limited data~\cite{jia2022visual,Qin2022MedicalIU}. The fundamental difference in technical routine has started reshaping the landscape of medical image analysis. 
Therefore, it is in urgent demand to set up datasets and benchmarks to promote innovation in this fast-marching research field and properly evaluate the performance gain and other cost-effective aspects. There are benchmarks~\cite{sun2020fewshot,shakeri2022fhist} for the few-shot learning tasks. Nonetheless, they focus more on each individual data modality and task. Here, we will instead promote the generalizability of the few-shot learning methods, i.e., strengthening their overall performance on various data modalities and tasks. 

\begin{figure}[t]
	\centering
        \begin{subfigure}[b]{0.18\linewidth}
             \centering
             \includegraphics[width=\linewidth]{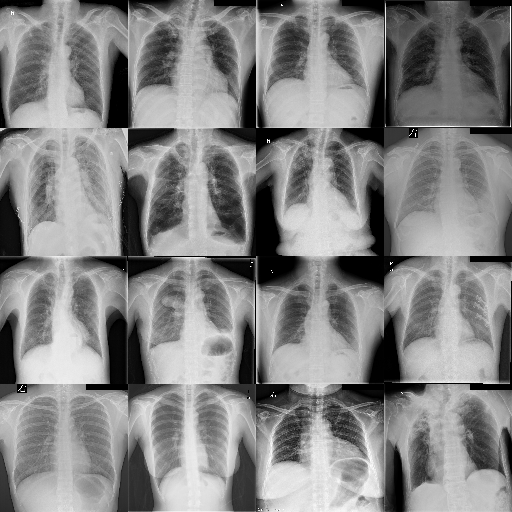}
             \caption{ChestDR}
             \label{fig:all-chest}
        \end{subfigure}
        \hfill
        \begin{subfigure}[b]{0.18\linewidth}
             \centering
             \includegraphics[width=\linewidth]{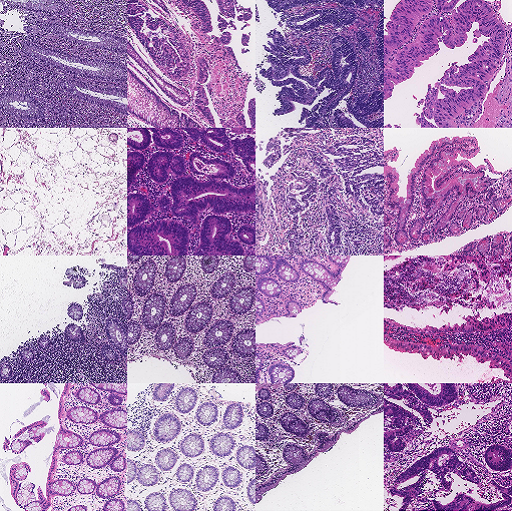}
             \caption{ColonPath}
             \label{fig:all-colon}
        \end{subfigure}
        \hfill
        \begin{subfigure}[b]{0.18\linewidth}
             \centering
             \includegraphics[width=\linewidth]{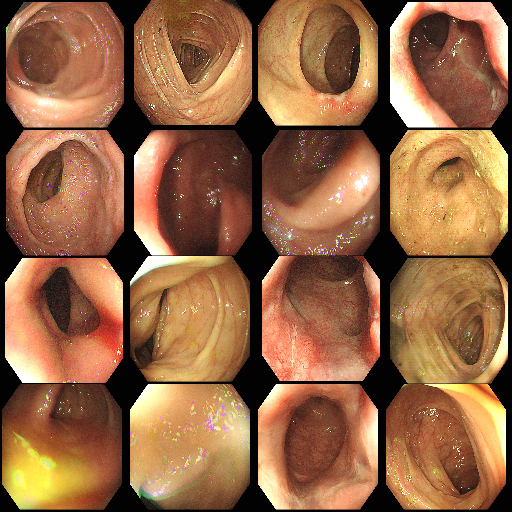}
             \caption{Endo}
             \label{fig:all-endo}
        \end{subfigure}
        \hfill
        \begin{subfigure}[b]{0.18\linewidth}
             \centering
             \includegraphics[width=\linewidth]{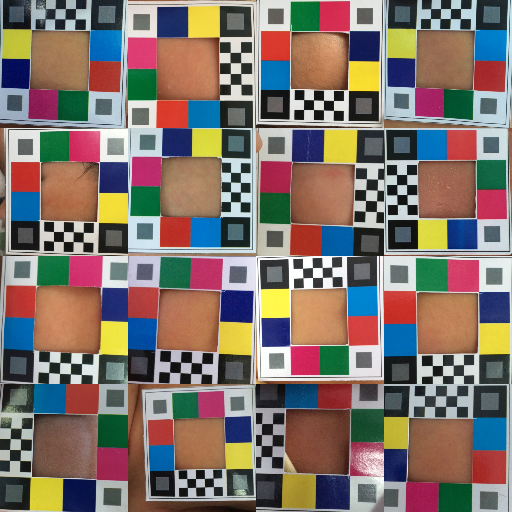}
             \caption{NeoJaundice}
             \label{fig:all-jaundice}
        \end{subfigure}
        \hfill
        \begin{subfigure}[b]{0.18\linewidth}
             \centering
             \includegraphics[width=\linewidth]{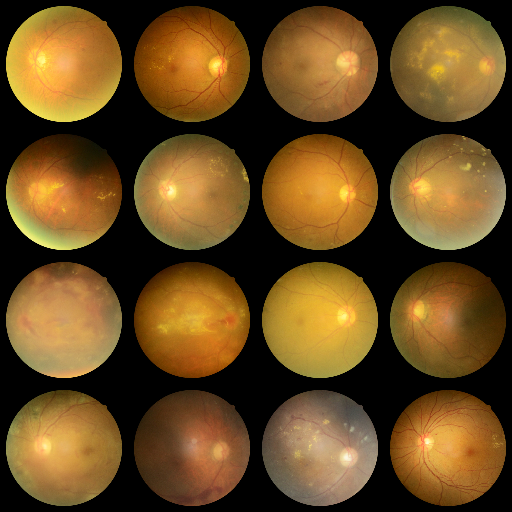}
             \caption{Retino}
             \label{fig:all-retino}
        \end{subfigure}
	\caption{Overview of the proposed learning-to-vote learning framework. 
	}
	\label{fig:data-overview}
\end{figure}

In this paper, we proposed a novel dataset, \textbf{\titleDataset }, with 22,349 images in total, which encapsulates five representative medical image classification tasks from real-world clinical daily routines. Fig.~\ref{fig:data-overview} presents sample images from each subset, and Table~\ref{tab:data-overview} shows the summary of data, including modality, number of samples, image size, classification tasks, and number of classes.  Different from many existing public datasets in the medical domain, e.g., Chest X-rays~\cite{Wang2017ChestXRay8HC,Irvin2019CheXpertAL,Johnson2019MIMICCXRAL}, MSD~\cite{Antonelli2021TheMS}, and HAM10000~\cite{Tschandl2018TheHD}, the proposed dataset and benchmark do not target advancing and evaluating the performance of each individual task with the conventional full-supervised training paradigm, which may require larger amount of data individually. Instead, we believe that this new dataset (as a union) provides valuable support to develop and \textbf{evaluate generalizable solutions of adapting foundation models} to a variety of medical downstream applications, e.g., using few samples as the prompts and the rest as testing standardly across all five tasks. In this study, we focus on 2D medical image classification as a start and cover the most common 2D medical imaging modalities. 3D data and other tasks, e.g., detection and segmentation, will be expanded and investigated in future work. 

\begin{table}[t]
\resizebox{\textwidth}{!}{
\begin{tabular}{cccccccc}
\hline
Name       & Modality     & Dimension & \# Sample & Image Size & Target                        & Task        & \# Class \\ \hline
ChestDR    & X-ray        & 2D        & 4,848     & 2953*2965    & Thoracic Abnormality          & Multi-label & 19       \\
ColonPath  & Pathology    & 2D        & 10,009    & 1024*1024  & Gastrointestinal Lesion & Binary & 2        \\
Endo       & Endoscopy    & 2D        & 3,865     & 1280*1024  & Colorectal Lesion             & Multi-label & 4        \\
NeoJaundice & Digital camera       & 2D        & 2,235     & 567*567    &  Neonatal Jaundice                 & Binary & 2        \\
Retino     & Retinography & 2D        & 1,392     & 2736*1824  & Diabetic Retinopathy     & Multi-class & 5        \\ \hline
\end{tabular}}
\caption{Data summary of \titleDataset }
\label{tab:data-overview}
\end{table}

\begin{figure}[t]
	\centering
        \begin{subfigure}[b]{0.45\linewidth}
             \centering
             \includegraphics[width=\linewidth]{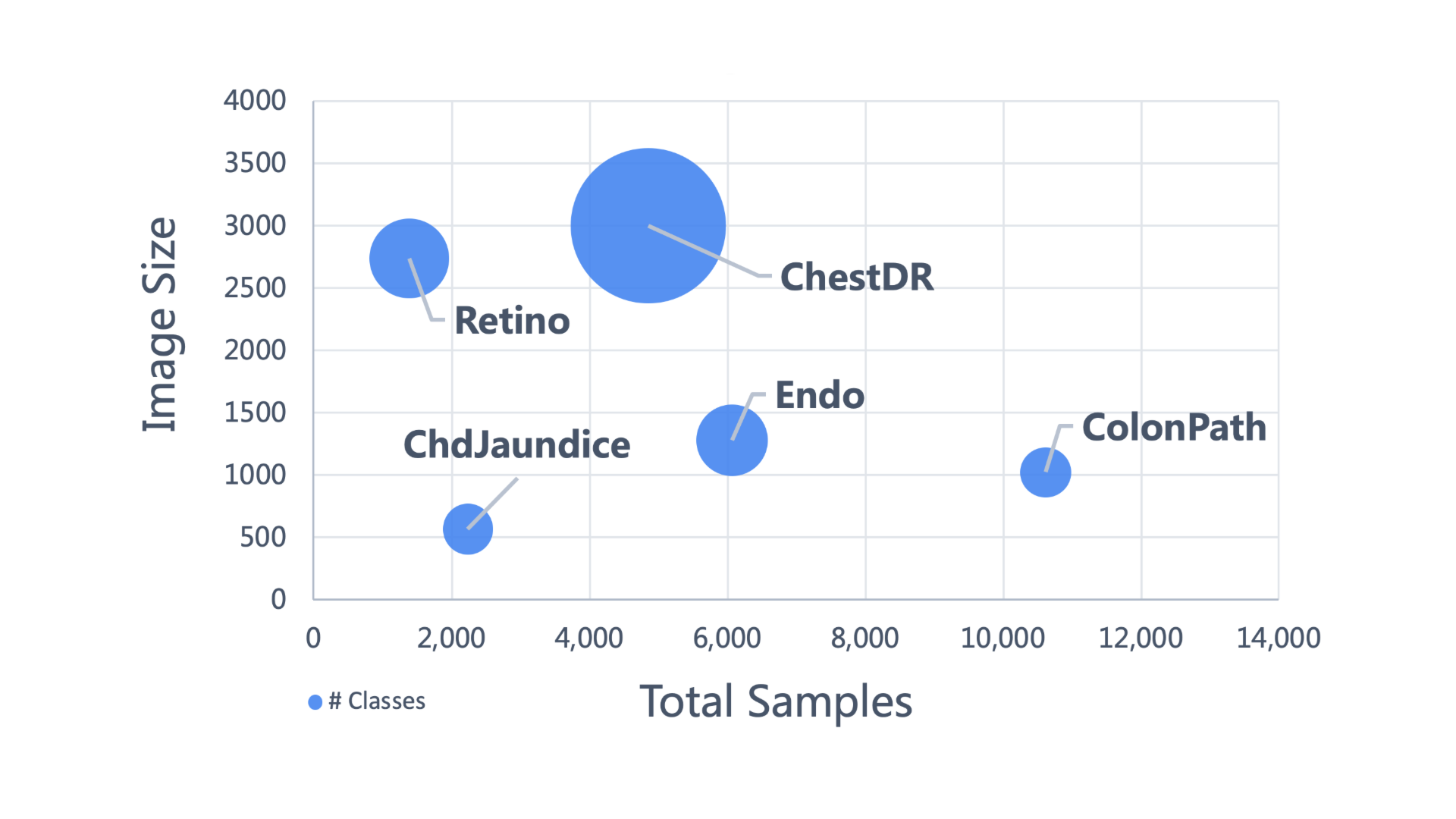}
             \caption{Diversified tasks and samples}
             \label{fig:data-overivew}
        \end{subfigure}
        \hfill
        \begin{subfigure}[b]{0.54\linewidth}
             \centering
             \includegraphics[width=\linewidth]{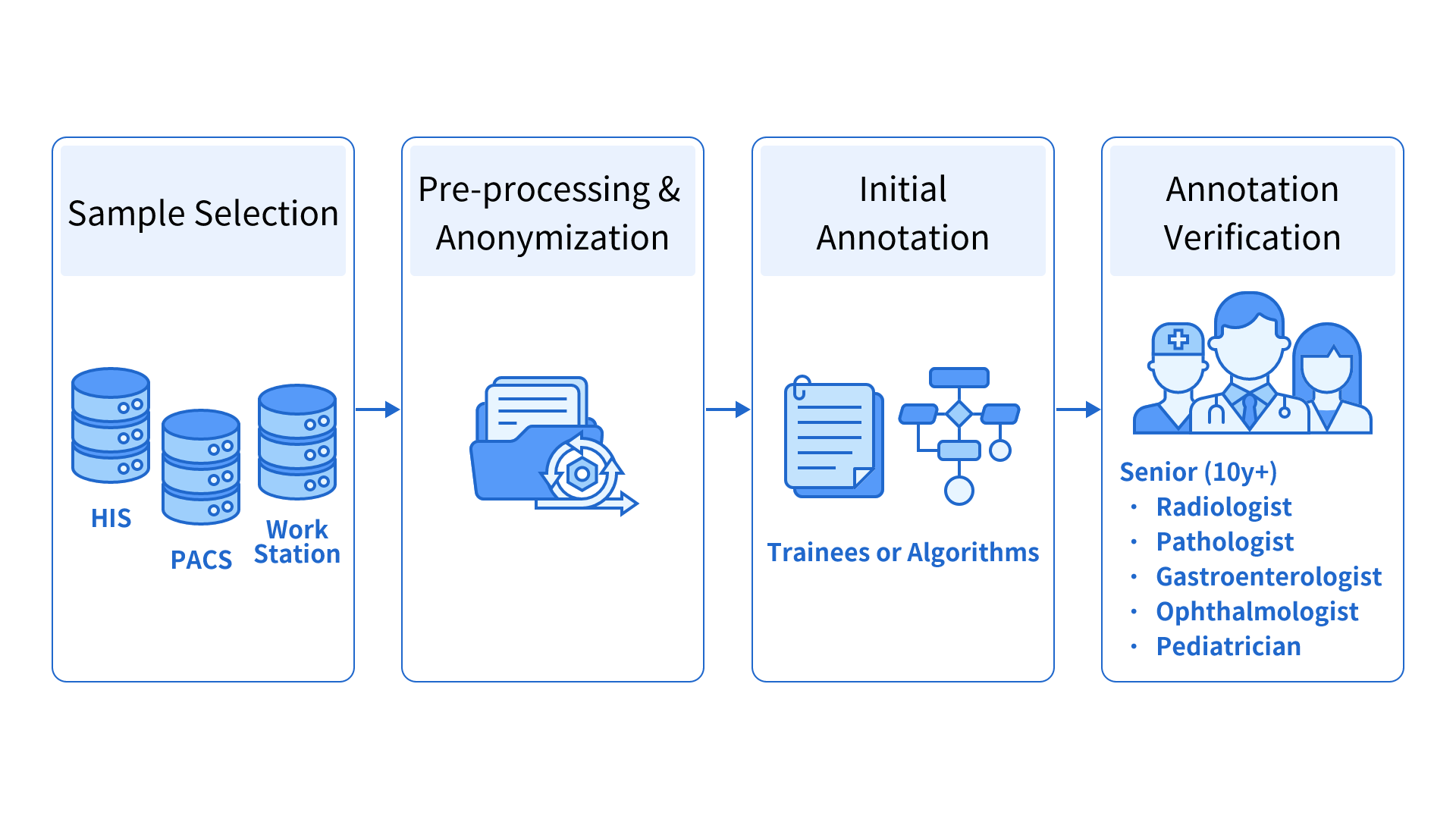}
             \caption{Data collection and annotation}
             \label{fig:data-pipeline}
        \end{subfigure}
	\caption{Overview of \titleDataset . 
	}
	\label{fig:overview}
\end{figure}

The proposed datasets target promoting the following aspects of foundation model adaptation approaches:

\begin{itemize}
  \item \textbf{Generalizability:} The proposed dataset has the capacity to examine the generalizability of the evaluated method from multiple perspectives. First, the benchmarked approach should achieve superior performance on all five prediction tasks, which are largely varied in data modality and image characteristics. Additionally, the composed five subsets of data are diversified in image sizes, data sample numbers, and classification tasks (e.g., multi-class, multi-label, and regression ones), as shown in Fig.~\ref{fig:data-overivew}.
  \item \textbf{Performance on Rare Diseases (Tail Classes):} The few-shot learning scheme fits perfectly for the long-tailed classification scenario, which often has only a few cases available for rare diseases in training. We will also face data scarcity in the testing phase, and separate evaluation metrics need to be recruited. The performance of algorithms on these tail classes can better reveal the power of pre-trained models and their adaptation techniques.
  \item \textbf{Prediction Accuracy and Adaptation Efficiency}: Besides evaluating the prediction accuracy of algorithms, we also pay attention to the efficiency of training (with fewer samples) in the cost of both data and computation. By combining both the accuracy and cost aspects in the evaluation metrics, we expect the advanced methods can further ease the effort of obtaining quality annotations and meanwhile lower the demand for computational resources.  
    
\end{itemize}

Illustratively, we present the benchmarking results of several common learning paradigms, e.g., fine-tuning and few-shot approaches. During the training phase, a small amount of randomly picked data (a few samples, i.e., 1, 5, and 10) are utilized for the initial training, and the rest of the dataset is employed for the validation. Approaches with advanced cross-domain knowledge transfer techniques are expected to achieve higher performance scores in such a setting. The final metrics are computed on an average of ten individual runs of the same testing process.  

\section*{Methods}



\subsection*{IRB Ethics Review and Exemption}
The presented retrospective research study has been reviewed by each involved institute individually, and patients consent to data sharing and the open publication of the data (otherwise waived as detailed below).  The ChestDR is approved by Fengcheng People's Hospital Ethics Committee (Ref. 2020 YiYanLunShen No.016) and Huanggang Hospital of Traditional Chinese Medicine Medical Research Ethics Committee (Ref. 2020 LunShen No.003). The NeoJaundice is approved by Xuzhou Central Hospital Ethics Committee (Ref. XZXYLQ-20180517-008), and patients' consent was signed by the guardians of the children. The Retino is approved by Shanghai Tenth People’s Hospital Ethics Committee (Ref. SHSY-IEC-4.1/20-154/01). The Endo is approved by Renji Hospital Ethics Committee. The committee reviewed and waived consent since the research was a retrospective study, and the risk of disclosing patient privacy via the studied snapshot images was minimized. The ColonPath is derived from part of the DigestPath 2019 challenge data, accessible via https://digestpath2019.grand-challenge.org/Dataset/, which was originally approved by the Histo Pathology Diagnostic Center Ethics Committee with patients' consent waived.  

\subsection*{Shared Pipeline for Data Collection and Annotation}
Fig. \ref{fig:data-pipeline} illustrates the general data sample collection and annotation pipeline. \titleDataset is composed of data with five different modalities in medical imaging, \ie, chest radiography, pathological images, endoscopy photos, dermatological images, and retinal images. The entire process consists of three major steps. First, the original data are listed and fetched from various systems, e.g., X-rays in the picture archiving and communication system (PACS), blood test results in Health Information System (HIS), endoscopy photos in the workstations, etc. Detailed processes are varied from modality to modality, which will be introduced in detail individually. Then, standardized anonymization of patient information (mainly the DICOM images) is performed before leaving the hospitals using the DICOM Anonymizer tool provided by the RSNA MIRC~\cite{rsna_mirc}. All image data are converted into 12-bit PNG images while the original image sizes are preserved. All image samples are manually examined to redact any privacy-related text or objects recorded in the images. Finally, a two-stage annotation process is conducted by first generating the initial labels, e.g., annotated by the medical trainees, blood test results extracted from the HIS, and grading prediction from a pre-trained model using public datasets. Senior professionals with over ten years of experience in their specialty, e.g., radiologist, pathologist, gastroenterologist, ophthalmologist, and pediatrician, verify the annotation for each image. In the following sections, we will discuss specific settings for each subset.

\subsubsection*{ChestDR: Thoracic Diseases Screening in Chest Radiography}

\begin{figure}[b]
	\centering
             \includegraphics[width=0.69\linewidth]{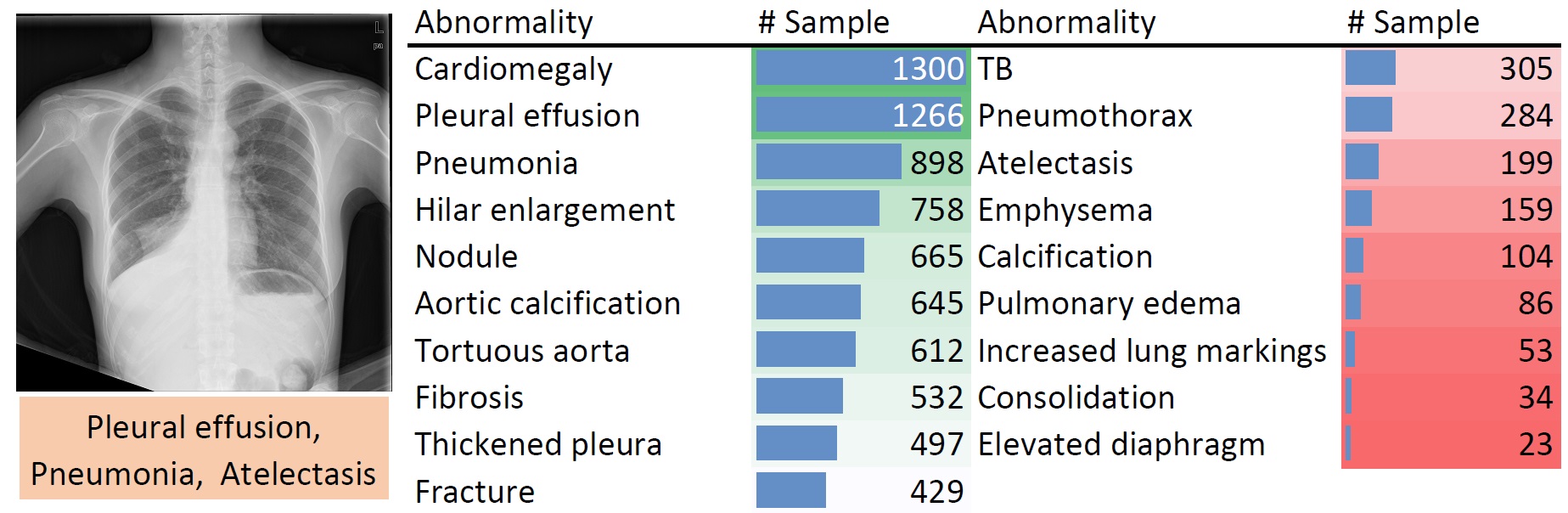}
	\caption{Data samples and case summary of ChestDR. }
	\label{fig:sample_chestdr}
\end{figure}

Chest X-ray is a regularly adopted imaging technique for daily clinical routine. Many thoracic diseases are reported, and further examinations are recommended for differential diagnoses. Due to the large amount and fast reporting requirements in certain emergency facilities, a swift screening and reporting of common thoracic diseases could largely improve the efficiency of the clinical process. Although a few chest x-ray datasets~\cite{Wang2017ChestXRay8HC,Irvin2019CheXpertAL,Johnson2019MIMICCXRAL} are now publicly available, images with quality annotations (preferably verified by radiologists) are still a desired resource for training and evaluating the models.  

A total of 4,848 frontal radiography images (from 4,848  patients) are provided in ChestDR, collected from two regional hospitals in Hubei and Jiangxi Province, China. A detailed distribution of 19 common thoracic diseases is presented in Fig.~\ref{fig:sample_chestdr}, which is sorted with the number of samples. Tail classes are highlighted in Red. Each PNG image is converted from the original DICOM files using the default window level and width (stored in the DICOM tags). The original image sizes are preserved. The initial disease labels are provided by a radiological resident (with the support of previously signed radiology reports) and then confirmed by a senior radiologist. 

\subsubsection*{ColonPath: Lesion Tissue Screening in Pathology Patches}
\begin{figure}[t]
	\centering
        \begin{subfigure}[b]{0.34\linewidth}
             \centering
             \includegraphics[width=\linewidth]{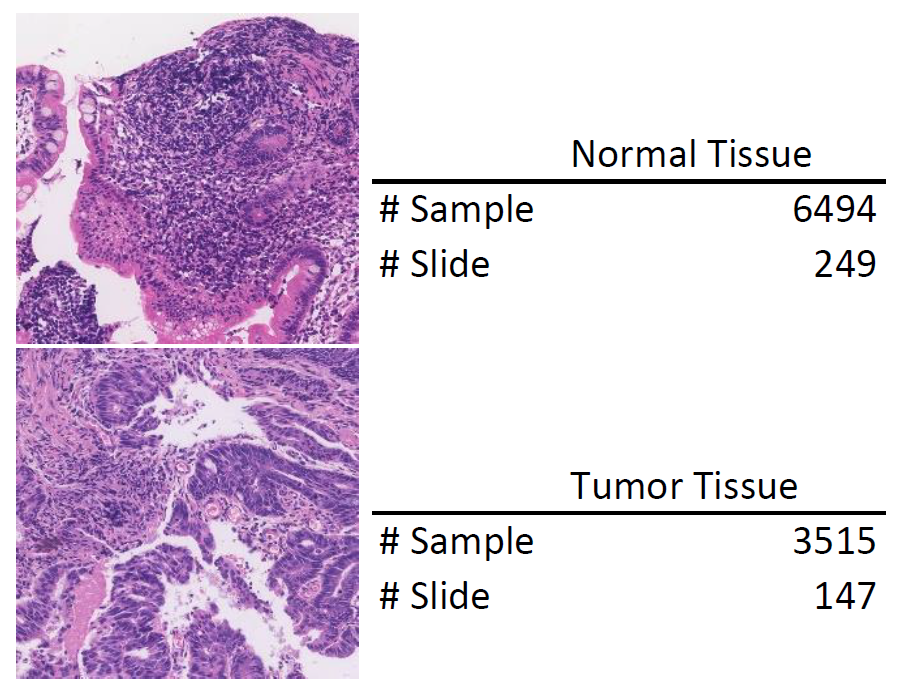}
             \caption{}
             \label{fig:sample-colonpath}
        \end{subfigure}
        \begin{subfigure}[b]{0.50\linewidth}
             \centering
             \includegraphics[width=\linewidth]{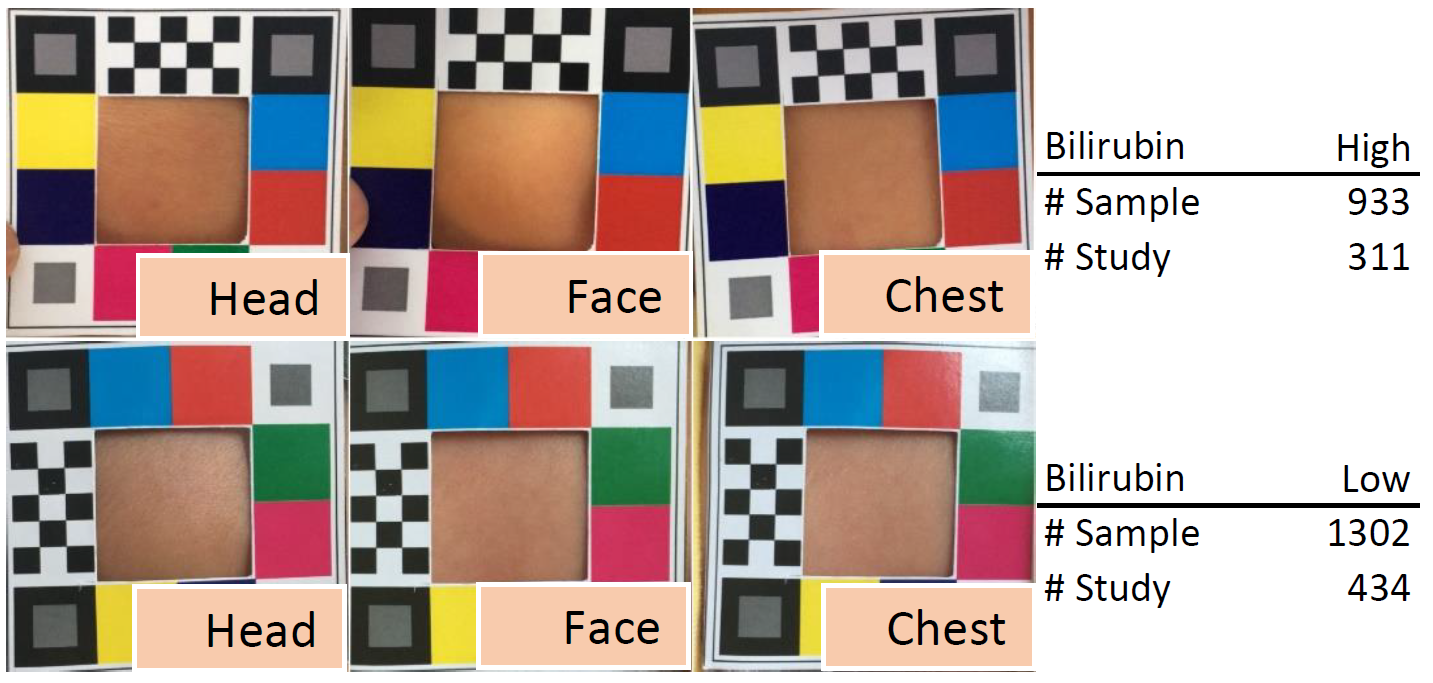}
             \caption{}
             \label{fig:sample-jaundice}
        \end{subfigure}
	\caption{Data samples and case summary of (a) ColonPath and (b) NeoJaundice.}
	\label{fig:colonpath-jaundice}
\end{figure}

Pathology examination can support detecting early-stage cancer cells in small tissue slices. In the pathologist's daily routine, they are required to look over several dozens of tissue slides, a tiresome and tedious job. In clinical diagnosis, quantifying cancer cells and regions is the primary goal for pathologists. The approaches for the classification of pathological tissue patches are desired to ease this process. They can help screen whether it exists regions of malignant cells in the entire slide in a sliding window manner.

The pathology whole slide image (WSI) is originally collected from the Histo Pathology Diagnostic Center, which is also published and utilized in the DigestPath Challenge 2019~\cite{da2022digestpath}. Only the data for the lesion segmentation tasks are employed in this study. All WSIs were acquired during 2017–2019 with hematoxylin and eosin (HE) stains and scanned using the KF- BIO FK-Pro-120 slide scanner. Subsequently, the WSIs were re-scaled to ×20 magnification with a pixel resolution of 0.475 $\mu$m.
Tissue patches are extracted from the WSI in a sliding window fashion with a fixed size of $1024\times1024$ and a stride of 768. A total of 396 patients’ 10,009 large tissue patches (with a uniform size of $1024\times1024$ ) of colonoscopy pathology examination will be available in ColonPath. Positive and negative patch samples (with and without the lesion tissue, computed based on the existing lesion region labels) are illustrated in Fig.~\ref{fig:sample-colonpath} along with the number of samples in each category.  The initial labels (whether containing lesion tissues) are provided by a trainee in the pathology specialty (with the support of computed labels) and then confirmed by a senior pathologist.

\subsubsection*{NeoJaundice: Neonatal Jaundice Evaluation in Skin Photos}

Jaundice commonly occurs in newborn infants. However, most jaundice is benign and does not require any interference. Conventionally, newborns must be monitored by taking a blood test to examine the bilirubin level. The potential toxicity of bilirubin might lead to severe hyperbilirubinemia and, in rare cases, acute bilirubin encephalopathy or kernicterus. Recent techniques utilized skin photos of three different parts of the infants, \ie, head, face, and chest, to estimate the total serum bilirubin in the blood so as to avoid the repeated invasive blood test for infants. 

A total of 745 infants’ 2,235 images (with an average size of $567\times567$ ) are collected in the NeoJaundice dataset from the Xuzhou central hospital. The initial binary labels are generated using the total serum bilirubin readings extracted from the hospital's health information system with a threshold of 12.9mg/dL and then confirmed by a senior experienced pediatrician. Samples of both low and high bilirubin levels are illustrated in Fig.~\ref{fig:sample-jaundice} along with the number of samples in each category. Three images are acquired for each infant on body skins of the head, face, and chest, using digital cameras.  The skin regions are surrounded by a standardized color card for color calibration purposes.

\subsubsection*{Endo: Lesion Classification in Colonoscopy Images}

\begin{figure}[t]
	\centering
        \begin{subfigure}[b]{0.4\linewidth}
             \centering
             \includegraphics[width=\linewidth]{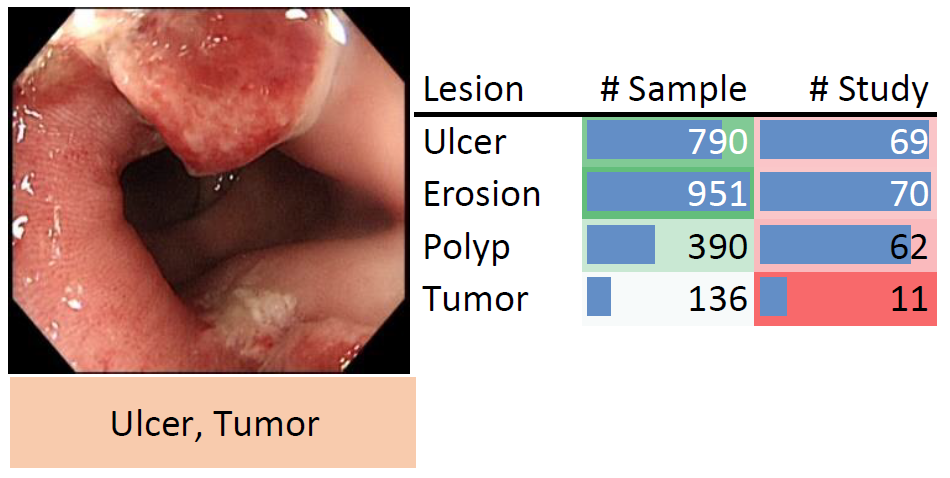}
             \caption{}
             \label{fig:sample-endo}
        \end{subfigure}
        \begin{subfigure}[b]{0.58\linewidth}
             \centering
             \includegraphics[width=\linewidth]{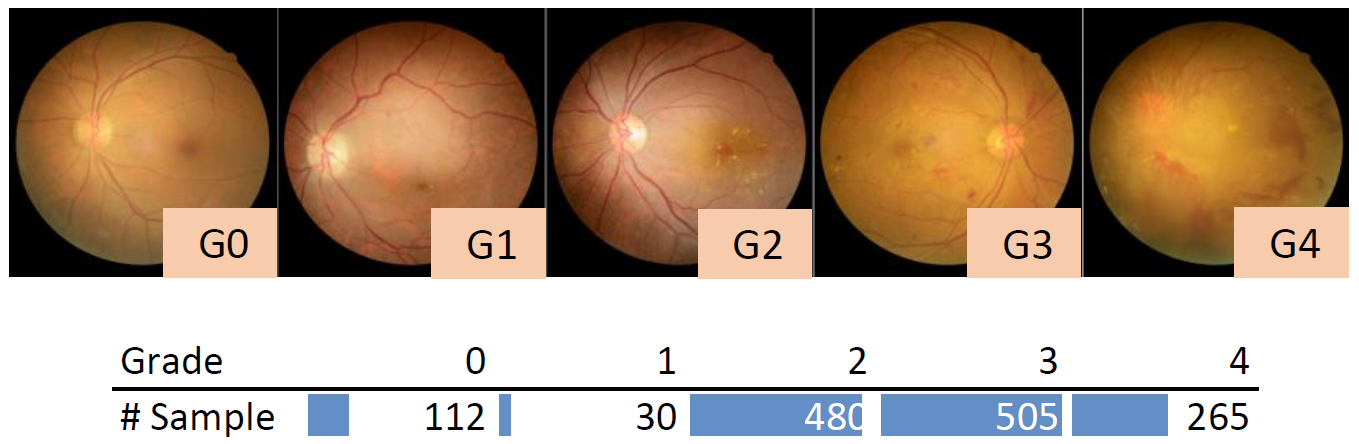}
             \caption{}
             \label{fig:sample-retino}
        \end{subfigure}
	\caption{Data samples and case summary of (a) Endo and (b) Retino.}
	\label{fig:endo_retino}
\end{figure}

Colorectal cancer is one of the most common and fatal cancers among men and women around the world. Abnormalities like polyps and ulcers are precursors to colorectal cancer and are often found in colonoscopy screening of people aged above 50. The risks largely increase along with aging. Colonoscopy is the gold standard for the detection and early diagnosis of such abnormalities with necessary biopsy on site, which could significantly affect the survival rate from colorectal cancer. Automatic detection of such lesions during the colonoscopy procedure could prevent missing lesions and ease the workload of gastroenterologists in colonoscopy. 

A total of 80 patients’ 3,865 images (with an average size of $1280\times1024$ ) recorded during the colonoscopy examination on the workstations in Renji Hospital are produced in the Endo dataset. Four types of lesions, \ie, ulcer, erosion, polyp, and tumor, are included, which are illustrated in Fig.~\ref{fig:sample-endo} along with the number of samples in each category. Non-relevant images are already excluded, while some noisy and degraded recordings remain to reflect the real-world data distribution. These noisy data are mainly caused by motions during the operation, which only occupy a small portion (<5\%) of the images and often are labeled without any of the target lesions. The initial labels of lesions are performed by a junior gastroenterologist (with the support of health records and reports) and then confirmed by a senior experienced gastroenterologist.

\subsubsection*{Retino: Diabetic Retinopathy Grading in Retina Images}


Diabetic retinopathy (DR) can lead to vision loss and blindness in patients with diabetes, mainly affecting the blood vessel in the retina. Therefore, it is important to have an exam of the retina each year for the early detection of DR.  Currently, DR grading requires a trained ophthalmologist to manually evaluate color fundus photos of the retina, which is time-consuming and may delay the treatment of patients. Automated screening of DR has long been recognized and desired.

A total of 1,392 patients’ fundus images (one from each patient with an average size of $2736\times1824$ ) from Shanghai Tenth People’s Hospital are included in the Retino dataset, which is extracted from the retinal imaging workstations after the examination. Images are captured by Canon nonmydriatic fundus cameras that mainly adopted the $45^\circ$ macula-centered imaging protocol. Samples of retina images in each of the five grades are illustrated in Fig.~\ref{fig:sample-retino} along with the number of samples in each grade. A DenseNet-121 (with ImageNet pre-trained model weights) is first fine-tuned using the dataset from Kaggle’s “Diabetic Retinopathy Detection” challenge and produced the prediction for each image. Then, an ophthalmologist with over ten years of experience examined again based on the automated generated prediction, i.e., the presence of diabetic retinopathy on a scale of 0 to 4 (0: No DR; 1: Mild; 2: Moderate; 3: Severe; 4: Proliferative DR). 







\section*{Data Records}

Each dataset in \titleDataset  consists of all image data in a ``images'' folder and associated image-level labels for each image in a CSV file.  Multi-label tasks (\ie, ChestDR and Endo) will have multiple columns with either 1 or 0 that represent the existence of corresponding disease patterns. Binary and multi-class classification tasks (\ie, ColonPath, NeoJaundice, and Retino) will have only a single label with the individual class number. The images are named differently across institutes, i.e., named with a random ID (ChestDR and NeoJaundice) and with a random ID together with the data of collection, not the examination (ColonPath, Endo, and Retino).


\section*{Technical Validation}

\begin{table}[t]
\centering
\resizebox{\textwidth}{!}{
\begin{tabular}{|c|cc|cc|cc|cc|cc|}
\hline
\multirow{2}{*}{\begin{tabular}[c]{@{}c@{}}Meta-baseline\\      Dense121(ImageNet, SL)~~~~\end{tabular}} & \multicolumn{2}{c|}{ChestDR}       & \multicolumn{2}{c|}{ColonPath}     & \multicolumn{2}{c|}{NeoJaundice}   & \multicolumn{2}{c|}{Endo}          & \multicolumn{2}{c|}{Retino}        \\ \cline{2-11} 
                                                                                                     & \multicolumn{1}{c|}{mAP}   & AUC   & \multicolumn{1}{c|}{Acc}   & AUC   & \multicolumn{1}{c|}{Acc}   & AUC   & \multicolumn{1}{c|}{mAP}   & AUC   & \multicolumn{1}{c|}{Acc}   & AUC   \\ \hline
1-shot                                                                                               & \multicolumn{1}{c|}{0.110} & 0.534 & \multicolumn{1}{c|}{0.682} & 0.798 & \multicolumn{1}{c|}{0.531} & 0.567 & \multicolumn{1}{c|}{0.202} & 0.657 & \multicolumn{1}{c|}{0.306} & 0.696 \\ \hline
5-shot                                                                                               & \multicolumn{1}{c|}{0.114} & 0.546 & \multicolumn{1}{c|}{0.731} & 0.856 & \multicolumn{1}{c|}{0.547} & 0.582 & \multicolumn{1}{c|}{0.205} & 0.647 & \multicolumn{1}{c|}{0.432} & 0.780 \\ \hline
10-shot                                                                                              & \multicolumn{1}{c|}{0.114} & 0.550 & \multicolumn{1}{c|}{0.735} & 0.863 & \multicolumn{1}{c|}{0.556} & 0.593 & \multicolumn{1}{c|}{0.208} & 0.658 & \multicolumn{1}{c|}{0.464} & 0.794 \\ \hline
\end{tabular}}
\hspace{2px}

\resizebox{\textwidth}{!}{
\begin{tabular}{|c|cc|cc|cc|cc|cc|}
\hline
\multirow{2}{*}{\begin{tabular}[c]{@{}c@{}}Meta-baseline\\      Swin-base(ImageNet, SL)~~~~\end{tabular}} & \multicolumn{2}{c|}{ChestDR}       & \multicolumn{2}{c|}{ColonPath}     & \multicolumn{2}{c|}{NeoJaundice}   & \multicolumn{2}{c|}{Endo}          & \multicolumn{2}{c|}{Retino}        \\ \cline{2-11} 
                                                                                                      & \multicolumn{1}{c|}{mAP}   & AUC   & \multicolumn{1}{c|}{Acc}   & AUC   & \multicolumn{1}{c|}{Acc}   & AUC   & \multicolumn{1}{c|}{mAP}   & AUC   & \multicolumn{1}{c|}{Acc}   & AUC   \\ \hline
1-shot                                                                                                & \multicolumn{1}{c|}{0.111} & 0.538 & \multicolumn{1}{c|}{0.647} & 0.742 & \multicolumn{1}{c|}{0.514} & 0.531 & \multicolumn{1}{c|}{0.155} & 0.521 & \multicolumn{1}{c|}{0.284} & 0.627 \\ \hline
5-shot                                                                                                & \multicolumn{1}{c|}{0.121} & 0.571 & \multicolumn{1}{c|}{0.709} & 0.861 & \multicolumn{1}{c|}{0.601} & 0.644 & \multicolumn{1}{c|}{0.167} & 0.531 & \multicolumn{1}{c|}{0.403} & 0.743 \\ \hline
10-shot                                                                                               & \multicolumn{1}{c|}{0.135} & 0.604 & \multicolumn{1}{c|}{0.762} & 0.884 & \multicolumn{1}{c|}{0.601} & 0.652 & \multicolumn{1}{c|}{0.168} & 0.546 & \multicolumn{1}{c|}{0.425} & 0.775 \\ \hline
\end{tabular}}
\hspace{2px}

\resizebox{\textwidth}{!}{
\begin{tabular}{|c|cc|cc|cc|cc|cc|}
\hline
\multirow{2}{*}{\begin{tabular}[c]{@{}c@{}}Meta-baseline\\      Swin-base(SimMIM, SSL)~~\end{tabular}} & \multicolumn{2}{c|}{ChestDR}       & \multicolumn{2}{c|}{ColonPath}     & \multicolumn{2}{c|}{NeoJaundice}   & \multicolumn{2}{c|}{Endo}          & \multicolumn{2}{c|}{Retino}        \\ \cline{2-11} 
                                                                                                     & \multicolumn{1}{c|}{mAP}   & AUC   & \multicolumn{1}{c|}{Acc}   & AUC   & \multicolumn{1}{c|}{Acc}   & AUC   & \multicolumn{1}{c|}{mAP}   & AUC   & \multicolumn{1}{c|}{Acc}   & AUC   \\ \hline
1-shot                                                                                               & \multicolumn{1}{c|}{0.108} & 0.533 & \multicolumn{1}{c|}{0.714} & 0.797 & \multicolumn{1}{c|}{0.536} & 0.564 & \multicolumn{1}{c|}{0.147} & 0.534 & \multicolumn{1}{c|}{0.327} & 0.647 \\ \hline
5-shot                                                                                               & \multicolumn{1}{c|}{0.125} & 0.579 & \multicolumn{1}{c|}{0.751} & 0.882 & \multicolumn{1}{c|}{0.574} & 0.611 & \multicolumn{1}{c|}{0.154} & 0.543 & \multicolumn{1}{c|}{0.392} & 0.739 \\ \hline
10-shot                                                                                              & \multicolumn{1}{c|}{0.131} & 0.595 & \multicolumn{1}{c|}{0.767} & 0.881 & \multicolumn{1}{c|}{0.597} & 0.642 & \multicolumn{1}{c|}{0.163} & 0.568 & \multicolumn{1}{c|}{0.458} & 0.797 \\ \hline
\end{tabular}}
\hspace{2px}

\resizebox{\textwidth}{!}{
\begin{tabular}{|c|cc|cc|cc|cc|cc|}
\hline
\multirow{2}{*}{\begin{tabular}[c]{@{}c@{}}VPT   \\      Swin-base (ImageNet, SSL)  \end{tabular}} & \multicolumn{2}{c|}{ChestDR}       & \multicolumn{2}{c|}{ColonPath}     & \multicolumn{2}{c|}{NeoJaundice}   & \multicolumn{2}{c|}{Endo}          & \multicolumn{2}{c|}{Retino}        \\ \cline{2-11} 
                                                                                             & \multicolumn{1}{c|}{mAP}   & AUC   & \multicolumn{1}{c|}{Acc}   & AUC   & \multicolumn{1}{c|}{Acc}   & AUC   & \multicolumn{1}{c|}{mAP}   & AUC   & \multicolumn{1}{c|}{Acc}   & AUC   \\ \hline
1-shot                                                                                       & \multicolumn{1}{c|}{0.131} & 0.565 & \multicolumn{1}{c|}{0.776} & 0.847 & \multicolumn{1}{c|}{0.584} & 0.559 & \multicolumn{1}{c|}{0.197} & 0.622 & \multicolumn{1}{c|}{0.415} & 0.645 \\ \hline
5-shot                                                                                       & \multicolumn{1}{c|}{0.171} & 0.648 & \multicolumn{1}{c|}{0.893} & 0.961 & \multicolumn{1}{c|}{0.644} & 0.686 & \multicolumn{1}{c|}{0.239} & 0.675 & \multicolumn{1}{c|}{0.456} & 0.727 \\ \hline
10-shot                                                                                      & \multicolumn{1}{c|}{0.190} & 0.667 & \multicolumn{1}{c|}{0.912} & 0.971 & \multicolumn{1}{c|}{0.667} & 0.727 & \multicolumn{1}{c|}{0.256} & 0.714 & \multicolumn{1}{c|}{0.527} & 0.752 \\ \hline
\end{tabular}}
\caption{Results of few-shot learning baseline on MedFMC}
\label{tab:base-few-shot}
\end{table}

\begin{table}[t]
\centering
\resizebox{0.9\textwidth}{!}{
\begin{tabular}{|c|ccccc|}
\hline
\textbf{ChestDR}              & \multicolumn{5}{c|}{Meta-Baseline ith Swin-base(ImageNet, SL)   and 10-shot}                                                                                              \\ \hline
Findings(Head)               & \multicolumn{1}{c|}{mAP}         & \multicolumn{1}{c|}{AUC}         & \multicolumn{1}{c|}{Findings(Tail)}                      & \multicolumn{1}{c|}{mAP}         & AUC         \\ \hline
cardiomegaly           & \multicolumn{1}{c|}{0.342±0.056} & \multicolumn{1}{c|}{0.577±0.060} & \multicolumn{1}{c|}{TB}                            & \multicolumn{1}{c|}{0.090±0.016} & 0.583±0.043 \\ \hline
pleural\_effusion      & \multicolumn{1}{c|}{0.411±0.063} & \multicolumn{1}{c|}{0.655±0.072} & \multicolumn{1}{c|}{pneumothorax}                  & \multicolumn{1}{c|}{0.109±0.020} & 0.640±0.037 \\ \hline
pneumonia              & \multicolumn{1}{c|}{0.220±0.041} & \multicolumn{1}{c|}{0.551±0.071} & \multicolumn{1}{c|}{atelectasis}                   & \multicolumn{1}{c|}{0.069±0.015} & 0.615±0.058 \\ \hline
hilar\_enlargement     & \multicolumn{1}{c|}{0.207±0.047} & \multicolumn{1}{c|}{0.562±0.075} & \multicolumn{1}{c|}{emphysema}                     & \multicolumn{1}{c|}{0.071±0.022} & 0.671±0.064 \\ \hline
nodule                 & \multicolumn{1}{c|}{0.144±0.016} & \multicolumn{1}{c|}{0.513±0.035} & \multicolumn{1}{c|}{calcification}                 & \multicolumn{1}{c|}{0.027±0.002} & 0.537±0.016 \\ \hline
aortic\_calcification  & \multicolumn{1}{c|}{0.191±0.025} & \multicolumn{1}{c|}{0.618±0.052} & \multicolumn{1}{c|}{pulmonary\_edema}              & \multicolumn{1}{c|}{0.104±0.009} & 0.811±0.015 \\ \hline
tortuous\_aorta        & \multicolumn{1}{c|}{0.208±0.037} & \multicolumn{1}{c|}{0.657±0.066} & \multicolumn{1}{c|}{increased\_lung\_marks}        & \multicolumn{1}{c|}{0.015±0.003} & 0.567±0.029 \\ \hline
fibrosis               & \multicolumn{1}{c|}{0.122±0.024} & \multicolumn{1}{c|}{0.509±0.071} & \multicolumn{1}{c|}{consolidation}                 & \multicolumn{1}{c|}{0.010±0.001} & 0.624±0.014 \\ \hline
thickness\_pleura      & \multicolumn{1}{c|}{0.102±0.012} & \multicolumn{1}{c|}{0.503±0.042} & \multicolumn{1}{c|}{elevated\_diaphragm}           & \multicolumn{1}{c|}{0.006±0.001} & 0.706±0.006 \\ \hline
fracture\_old          & \multicolumn{1}{c|}{0.119±0.014} & \multicolumn{1}{c|}{0.585±0.036} & \multicolumn{1}{c|}{\textbf{Avg(Tail)}}        & \multicolumn{1}{c|}{0.056±0.010} & 0.639±0.031 \\ \hline
\textbf{Avg(Head)} & \multicolumn{1}{c|}{0.205±0.034} & \multicolumn{1}{c|}{0.573±0.058} & \multicolumn{1}{c|}{\textbf{Avg(All classes)}} & \multicolumn{1}{c|}{0.135±0.024} & 0.604±0.045 \\ \hline
\end{tabular}}
\hspace{4px}

\resizebox{0.90\textwidth}{!}{
\begin{tabular}{|c|cc|c|cc|}
\hline
\multicolumn{1}{|c|}{\textbf{Endo}}                                                                                & \multicolumn{2}{c|}{}                                    & \multicolumn{1}{c|}{\textbf{Retino}}                                                                                                                                                   & \multicolumn{2}{c|}{}                          \\ \hline
\multicolumn{1}{|c|}{~~~~~~~Lesion Types~~~~~~} & \multicolumn{1}{c|}{mAP}         & AUC                   & \multicolumn{1}{c|}{~~~~~~~~~~~~Grade~~~~~~~~~~~~~~} & \multicolumn{1}{c|}{Acc}         & AUC         \\ \hline
ulcer                                                                                                              & \multicolumn{1}{c|}{0.251±0.028} & 0.586±0.035           & 0                                                                                                                                                                                      & \multicolumn{1}{c|}{0.567±0.056} & 0.865±0.036 \\ \hline
erosion                                                                                                            & \multicolumn{1}{c|}{0.320±0.056} & 0.577±0.040           & 1                                                                                                                                                                                      & \multicolumn{1}{c|}{0.665±0.258} & 0.919±0.028 \\ \hline
polyp                                                                                                              & \multicolumn{1}{c|}{0.093±0.010} & 0.544±0.031           & 2                                                                                                                                                                                      & \multicolumn{1}{c|}{0.341±0.065} & 0.581±0.086 \\ \hline
tumor                                                                                                              & \multicolumn{1}{c|}{0.009±0.001} & 0.478±0.028           & 3                                                                                                                                                                                      & \multicolumn{1}{c|}{0.525±0.102} & 0.729±0.030 \\ \hline
\textbf{Avg}                                                                                                   & \multicolumn{1}{c|}{0.168±0.015} & 0.546±0.015           & 4                                                                                                                                                                                      & \multicolumn{1}{c|}{0.416±0.044} & 0.779±0.017 \\ \hline
\multicolumn{1}{|l|}{}                                                                                             & \multicolumn{1}{l|}{}            & \multicolumn{1}{l|}{} & \textbf{Avg}                                                                                                                                                                       & \multicolumn{1}{c|}{0.425±0.046} & 0.775±0.023 \\ \hline
\end{tabular}}
\caption{Results of sub-classes with meta-baseline and 10-shot patient data.}
\label{tab:subclass}
\end{table}

\begin{table}[t]
\centering
\resizebox{\textwidth}{!}{
\begin{tabular}{|l|cc|cc|cc|cc|cc|cc|}
\hline
\multicolumn{1}{|c|}{\multirow{2}{*}{\begin{tabular}[c]{@{}c@{}}Fine-tuning\\      10-shot\end{tabular}}} & \multicolumn{2}{c|}{ChestDR (Head)}                   & \multicolumn{2}{c|}{ChestDR (Tail)}                   & \multicolumn{2}{c|}{ColonPath}                        & \multicolumn{2}{c|}{NeoJaundice}                      & \multicolumn{2}{c|}{Endo}                             & \multicolumn{2}{c|}{Retino}                           \\ \cline{2-13} 
\multicolumn{1}{|c|}{}                                                                                    & \multicolumn{1}{c|}{mAP}   & \multicolumn{1}{c|}{AUC} & \multicolumn{1}{c|}{mAP}   & \multicolumn{1}{c|}{AUC} & \multicolumn{1}{c|}{Acc}   & \multicolumn{1}{c|}{AUC} & \multicolumn{1}{c|}{Acc}   & \multicolumn{1}{c|}{AUC} & \multicolumn{1}{c|}{mAP}   & \multicolumn{1}{c|}{AUC} & \multicolumn{1}{c|}{Acc}   & \multicolumn{1}{c|}{AUC} \\ \hline
DenseNet-121                                                                                              & \multicolumn{1}{l|}{0.188} & 0.568                    & \multicolumn{1}{l|}{0.061} & 0.647                    & \multicolumn{1}{l|}{0.755} & 0.883                    & \multicolumn{1}{l|}{0.610} & 0.637                    & \multicolumn{1}{l|}{0.200} & 0.622                    & \multicolumn{1}{l|}{0.359} & 0.753                    \\ \hline
Efficient-b4                                                                                              & \multicolumn{1}{l|}{0.270} & 0.654                    & \multicolumn{1}{l|}{0.075} & 0.664                    & \multicolumn{1}{l|}{0.820} & 0.901                    & \multicolumn{1}{l|}{0.595} & 0.641                    & \multicolumn{1}{l|}{0.233} & 0.673                    & \multicolumn{1}{l|}{0.576} & 0.853                    \\ \hline
Swin-base ~~~~~~~~~~~~~~~                                                                                                & \multicolumn{1}{l|}{0.247} & 0.631                    & \multicolumn{1}{l|}{0.064} & 0.611                    & \multicolumn{1}{l|}{0.792} & 0.893                    & \multicolumn{1}{l|}{0.652} & 0.687                    & \multicolumn{1}{l|}{0.233} & 0.688                    & \multicolumn{1}{l|}{0.473} & 0.757                    \\ \hline
\end{tabular}}
\hspace{2px}

\resizebox{\textwidth}{!}{
\begin{tabular}{|l|cc|cc|cc|cc|cc|cc|}
\hline
\multicolumn{1}{|c|}{\multirow{2}{*}{\begin{tabular}[c]{@{}c@{}}Fine-tuning\\      20\%\end{tabular}}} & \multicolumn{2}{c|}{ChestDR (Head)} & \multicolumn{2}{c|}{ChestDR (Tail)} & \multicolumn{2}{c|}{ColonPath}     & \multicolumn{2}{c|}{NeoJaundice}   & \multicolumn{2}{c|}{Endo}          & \multicolumn{2}{c|}{Retino}        \\ \cline{2-13} 
\multicolumn{1}{|c|}{}                                                                                 & \multicolumn{1}{c|}{mAP}    & AUC   & \multicolumn{1}{c|}{mAP}    & AUC   & \multicolumn{1}{c|}{Acc}   & AUC   & \multicolumn{1}{c|}{Acc}   & AUC   & \multicolumn{1}{c|}{mAP}   & AUC   & \multicolumn{1}{c|}{Acc}   & AUC   \\ \hline
DenseNet-121                                                                                           & \multicolumn{1}{c|}{0.348}  & 0.745 & \multicolumn{1}{c|}{0.130}  & 0.761 & \multicolumn{1}{c|}{0.961} & 0.991 & \multicolumn{1}{c|}{0.742} & 0.814 & \multicolumn{1}{c|}{0.414} & 0.802 & \multicolumn{1}{c|}{0.699} & 0.910 \\ \hline
Efficient-b4                                                                                           & \multicolumn{1}{c|}{0.377}  & 0.744 & \multicolumn{1}{c|}{0.196}  & 0.804 & \multicolumn{1}{c|}{0.970} & 0.996 & \multicolumn{1}{c|}{0.752} & 0.829 & \multicolumn{1}{c|}{0.370} & 0.782 & \multicolumn{1}{c|}{0.696} & 0.912 \\ \hline
Swin-base                                                                                              & \multicolumn{1}{c|}{0.415}  & 0.782 & \multicolumn{1}{c|}{0.194}  & 0.789 & \multicolumn{1}{c|}{0.956} & 0.982 & \multicolumn{1}{c|}{0.716} & 0.794 & \multicolumn{1}{c|}{0.414} & 0.794 & \multicolumn{1}{c|}{0.787} & 0.948 \\ \hline
\begin{tabular}[c]{@{}l@{}}Swin-base   (No Aug)\end{tabular}                                   & \multicolumn{1}{c|}{0.407}  & 0.763 & \multicolumn{1}{c|}{0.142}  & 0.75  & \multicolumn{1}{c|}{0.951} & 0.986 & \multicolumn{1}{c|}{0.727} & 0.796 & \multicolumn{1}{c|}{0.419} & 0.795 & \multicolumn{1}{c|}{0.758} & 0.925 \\ \hline
\end{tabular}}
\caption{Results of transfer learning baseline on MedFMC with 10-shot and 20\% patient data.}
\label{tab:base-finetune}
\end{table}

\textbf{Dataset Partition:} Each image subset is divided into two parts: the few-shot pool and testing subsets. The few-shot pool consists of samples with about 20\% randomly selected patients, and the count of each class must be larger than 10. The remaining samples are used for testing. In transfer learning, we use all the images in the few-shot pool for training and validate the deep-learning-based classifier models using testing. In the few-shot setting, we randomly picked images of 1, 5, and 10 patients for each class from the few-shot pool to build the support set, and the testing subset is reserved for the model evaluation. We provide the data list of the few-shot pool and testing set together with sample lists of few-shot images in the repository (see the Code availability section).

\subsection*{Few-shot Learning Baseline}
In the experiment, we employ two few-shot baseline methods, i.e., Meta-Baseline~\cite{Chen2020MetaBaselineES} and Visual Prompt Tuning (VPT)~\cite{jia2022visual}. Meta-Baseline~\cite{Chen2020MetaBaselineES} is chosen here as a classic few-shot method to evaluate across all five datasets. The input images are converted to the embedding features via three backbone networks and pre-trained model settings, including DenseNet 121 layers (Dense121) with ImageNet pre-trained weights in supervised learning (SL) and a Swin Transformer (Swin-base) with pre-trained weights from both fully-supervised and self-supervised learning (SSL) schemes (SimMIM~\cite{xie2021simmim}, a form of Masked Auto-Encoder~\cite{he2022masked}). Settings are specified when reporting the performance as shown in the left column of Table~\ref{tab:base-few-shot}. We cluster the class centers in the support set using the extracted features and compute the cosine similarities between one image in the testing set and the class centers to determine the category. 
Additionally, we include VPT as an advanced method in training visual prompts for the few-shot classification tasks. In this case, a vanilla pre-trained model from the Swin-transformer repository (pre-trained on ImageNet21K and finetuned on ImageNet1k) is utilized to initialize the VPT-based few-shot tuning. 
We repeat the experiment 10 times (randomly picking few-shot samples) on the five medical image datasets and report the averaged testing results. 

\subsection*{Transfer Learning Baseline}
We run the fine-tuning experiments using three representative networks, including DenseNet, EfficientNet, and Swin Transformer, on the five medical image datasets. The Swin transformer model is pre-trained on ImageNet21K with self-supervised learning and then finetuned on ImageNet1k with labels. The others are also pre-trained using ImageNet but with supervised learning. In our experiments, the fine-tuning is performed as linear probing, i.e., only tuning the classifier (fully connected) layers since the parameters in the representation layers are also frozen for the few-shot baseline methods. We also experimented with finetuning the entire network, which could generally improve the performance by 1-2\% in accuracy. During the training and inference stage, all the input images are padded and rescaled to 384*384 pixels. Common data augmentation tricks, i.e., random crop, resize, and horizontal flip, are adopted. The cross-entropy loss is employed as the loss function for the multi-class classification of three datasets, including ColonPath, NeoJaundice, and Retino, while the binary cross-entropy loss is computed for the multi-label classification of the remaining two datasets, i.e., ChestDR and Endo. The model parameters (except the fully connected classifier layer) are initialized by the ImageNet pre-trained model weights and frozen during the tuning. SGD optimizers with initial learning rates of 0.002 and 0.01 are applied for the model training of DenseNet and EfficientNet, respectively. The Swin transformer model is optimized by AdamW with an initial learning rate of 0.001. We trained these classification models on a single NVIDIA A100 for 20 epochs at a batch size of 8, using the framework of MMClassification~\cite{openmmclass}.

\subsection*{Evaluation metrics}
To evaluate the performance of transfer learning and few-shot learning baseline experimental results, we compute the overall accuracy (Acc) and area under the receiver operating characteristic curve (AUC) for the multi-class classification tasks in the datasets of ColonPath, NeoJaundice, and Retino, and the mean average precision (mAP) and AUC for the multi-label classification tasks in the datasets of ChestDR and Endo.
Accuracy reflects the overall correct predictions among all the test images. The predicted label is determined with the maximum softmax outputs in the multi-class classification task.
AUC is computed for each class to measure the capability of distinguishing between positive and negative classes at various threshold settings.
The AP is the weighted average of precisions, while the mAP for all samples is the mean value of the AP scores for each class.

\subsection*{Benchmarking Results}
\noindent\textbf{Results of few-shot baselines:} The classification performance of few-shot baselines on each dataset is shown in Tables~\ref{tab:base-few-shot}. 
More data can often provide better support for distinguishing the representations of testing data, but it comes with a higher data demand and more extensive computation cost. The classification performance on five datasets varies significantly, which indeed indicates the diverse task difficulty. The Meta-baseline also performs better on parts of the five sets and also has mixed results for multi-class and multi-label classifications. VPT clearly achieves the best overall performance considering additional tuning parameters (visual prompts) and a network fine-tuning process included in the approach. Regarding the network backbone, advanced architectures, e.g., Swin-transformer, does not always produce superior performance over convolutional neural network counterparts when using the same ImageNet pre-trained model (via either supervised learning-based or self-supervised learning-based schemes). Furthermore, the detailed performance of each disease/lesion class for the three multi-label and multi-class classification tasks are illustrated in Table~\ref{tab:subclass}. Especially the results for thoracic diseases classification are listed for head and tail classes separately. Higher or equivalent AUCs for these rare classes (tail ones) are achieved, which indicates that few-shot methods can benefit the classification of rare classes more than the common learning paradigms.   

\noindent\textbf{Choices of few-shot samples:} Since we repeat the experiment 10 times (randomly picking few-shot samples) on the five medical image datasets. The choice can affect the classification performance of the averaged testing results. We list the STD in addition to the mean accuracy and AUC, as shown in Table~\ref{tab:subclass}. The variances of accuracy and AUCs are often fluctuant less than 5\% in the example 10-shot setting.  

\noindent\textbf{Results of fine-tuning baselines:} Table~\ref{tab:base-finetune} shows the results of fine-tuning-based classification frameworks with all 20\% patient data from the few-shot pool and with 10-shot sample data individually. We further list the results of head and tail classes (two columns as shown in Fig. \ref{fig:sample_chestdr}). There is still quite a gap between the classification accuracies for these two groups of methods when all 20\% of data in the few-shot pool are utilized in fine-tuning, which is reasonable, considering more training samples are utilized. Nonetheless, the fine-tuning performance decrease to an equivalent level to the few-shot learning paradigms when only ten patients' data are employed. When there is a scarcity of sample data, it is highly advantageous to utilize few-shot-based techniques. We do not show the results of fine-tuning using fewer (1 and 5) samples since we find the training hard to accomplish (either overfitting or underfitting) using very few data points, which reveals a critical limitation for the fine-tuning-based methods. Moreover, we provided the finetuning results with and without data augmentation at the bottom of Table~\ref{tab:base-finetune}. The difference between them is rather marginal. 

\section*{Usage Notes}

The \titleDataset  Dataset will be published via figshare (upon acceptance) . The provided dataset is publicly available under the Creative Commons Zero (CC0) Attribution. Please note the presented datasets are not intended for the development of diagnosis-oriented algorithms and models. It should also not be utilized as the sole base of the clinical evaluation for each classification task. 



\section*{Code availability}
The code repository of the presented few-shot methods can be accessed via GitHub (upon acceptance). 
No custom code was used to generate or process the data described in the manuscript. 

\bibliography{medfs5}


\section*{Acknowledgements} 

J.S. is supported by grants from Shanghai Science and Technology Innovation Initiative (21SQBS02302), and Cultivated Funding for Clinical Research Innovation, Ren Ji Hospital, Shanghai Jiao Tong University School of Medicine [RJPY-LX-004]. Q.D. is supported by Shanghai Municipal Science and Technology Key Project (Grant No. 20511100302).

\section*{Author contributions statement}

D.W. and X.W. conceptualized and compiled the dataset, created annotation protocols, and wrote most of the manuscript. L.W. and M.L. performed the technical validation. Q.D., X.L., X.G., and J.S. contributed to dataset curation and annotation. Q.D., T.S., and J.H. contributed to the dataset curation. J.Z., K.L., Y.Q., and S.Z. provided important scientific input and contributed to the writing of the manuscript. All authors read and approved the final version of the manuscript.

\section*{Competing interests}

The authors declare no competing interests

\end{document}